# Automated Testing for Deep Learning Systems with Differential Behavior Criteria

Yuan Gao[1], Yiqiang Han[2]

*Abstract*— **In this work, we conducted a study on building an automated testing system for deep learning systems based on differential behavior criteria. The automated testing goals were achieved by jointly optimizing two objective functions: maximizing differential behaviors from models under testing and maximizing neuron coverage. By observing differential behaviors from three pre-trained models during each testing iteration, the input image that triggered erroneous feedback was registered as a corner-case. The generated corner-cases can be used to examine the robustness of DNNs and consequently improve model accuracy. A project called DeepXplore was also used as a baseline model. After we fully implemented and optimized the baseline system, we explored its application as an augmenting training dataset with newly generated corner cases. With the GTRSB dataset, by retraining the model based on automated generated corner cases, the accuracy of three generic models increased by 259.2%, 53.6%, and 58.3%, respectively. Further, to extend the capability of automated testing, we explored other approaches based on differential behavior criteria to generate photo-realistic images for deep learning systems. One approach was to apply various transformations to the seed images for the deep learning framework. The other approach was to utilize the Generative Adversarial Networks (GAN) technique, which was implemented on MNIST and Driving datasets. The style transferring capability has been observed very effective in adding additional visual effects, replacing image elements, and style-shifting (virtual image to real images). The GAN-based testing sample generation system was shown to be the next frontier for automated testing for deep learning systems.**

## I. INTRODUCTION

In recent rapid developments, the Deep Learning (DL) system has achieved exceptional prediction accuracy in a growing number of tasks. Still, the complexity of the system itself has led to many problems, such as testing difficulty, interpretability, and other issues. These have become the major impediment during the deployment of DL systems in safety-crucial applications (e.g., autonomous driving, medical applications). A testing system that can automatically detect these potential issues during model validation and verification processes is highly desirable. In this paper, we propose such a testing tool based on differential behavior criteria to facilitate the training of the DL systems and also system security testing.

## II. RESEARCH OBJECTIVES

Comparing DL systems with traditional software systems, the most different point is the specific logic of the system. For the traditional software systems, the developers can directly define and set the decision logic. However, for the DL systems, the DNN components will automatically learn their decision logical rules from training data. The developers can modify three parts, namely, training data, feature selection, and algorithm tuning, to indirectly influence the decision logic learned by a DNN.

Due to the unknown DNNs' decision logic, we cannot easily identify erroneous behaviors of DNNs, so how to test and fix the erroneous behaviors of DNNs is especially crucial in safety-critical scenarios. Therefore, the goal of this study is to automatically generate numerous corner inputs that can reach different parts of the neural network, i.e., different neurons, and identify different erroneous behaviors of the DL systems.

There are multiple essential features related to this goal. The first feature is diversity. The testing system is expected to reach as many neurons as possible so that it can reveal different erroneous behaviors as many as possible. Then we need efficiency. The test inputs should be generated in a fast way for any tested system, so the testing method can more time-efficient. In addition, the testing method needs to be automated, so that no manual labeling or checking is necessary. Finally, all generated inputs need to be realistic. Therefore, they represent potential errors or attacks related to DNNs in the real world.

## III. METHODOLOGY

To achieve the goal mentioned above, we are solving a joint optimization problem guided by gradient ascent method. Two objectives (loss functions) have been identified by Ref. [1], namely:

1. Maximizing differential behaviors
2. Maximizing neuron coverage

where, Objective 1 can be expressed as:

$$obj_1(x) = (\Sigma_{i \neq j} F_i(x)[c] - \lambda_1 F_j(x)[c]) \qquad (1)$$

and Objective 2 can be found as:

$$obj_2(x) = f_n(x) \; s.t. \; f_n(x) > t \qquad (2)$$

Jointly, the optimization problem can be solved together in the form below using gradient ascent:

---

[1] Software Engineer III, GEICO Headquarter, Washington D.C.
[2] Research Assistant Professor, Clemson University, yiqianh@g.clemson.edu

$$obj_{joint} = \left(\Sigma_{i \neq j} F_i(x)[c] - \lambda_1 F_j(x)[c]\right) + \lambda_2 f_n(x) \quad (3)$$

The detailed explanations on important parameters are shown in following subsections.

*A. Objective 1: Maximizing differential behaviors*

A successful generation of a corner case is determined by observing the differential outputs from several similar DNNs. Here, multiple DNNs are used to provide cross-referencing results, for instance, in Ref. [1], 15 state-of-art (pre-trained) DNNs on five large datasets were used. The testing cases are generated based on a group of seed samples with small modifications guided by gradients between layers. The generation of such corner cases is obtained if at least one erroneous outputs are found that are different from others. The weight of such abnormal case is governed by an empirical hyperparameter. In this study, we will explore the effect of varying values of hyperparameters. Overall, we would like to find the testing case that maximizes this objective value. In this way, we can find DNN errors without the need for manual labels.

*B. Objective 2: Maximizing neuron coverage*

To achieve a maximized branch coverage of the testing cases, the authors in Ref. [1] proposed a separate test objective, which is called "neuron coverage." The assumption is that, in each of the DL layers, each neuron represents a higher-level abstraction of the object features, such as the color of the object, an edge/shape/part of the object, etc. This has also been studied in DNN visualization researches, where the neurons on each layer can also be used for layer-wise feature detection, such as Ref. [2]. A DNN system trained on comprehensive testing cases that can ensure the coverage of activation of all neurons can maximize the possibility of avoiding the erroneous outputs due to corner cases. The authors pointed out that we need a parameter to measure the completeness (coverage) of the testing cases, which is the so-called neuron coverage. The neuron coverage is defined as the ratio between the number of neurons that have been activated so far and the number of the total neurons in the DNN system. The corresponding test objective is then to maximize the coverage ratio during the testing case generation by increasing as many $f_n(x)$ above the activation threshold $t$ as possible shown in Equation (2). As shown in Equation (3), the balance between the two objectives is set using a hyperparameter $\lambda_2$, which is again, empirically determined.

*C. Joint training*

By introducing randomly initialized noise and distortion, we continuously observe the two objectives through the training process. We then feed the training samples to at least three custom trained DL systems and pick the training samples that triggers the differential behaviors in the systems. We have specifically studied the optimization of the hyperparameters of the framework and explored the application of augmenting training data with error-inducing samples generated by the testing system. We then re-trained DNNs to improve robustness against erroneous behaviors and malicious inputs.

*D. Domain-Specific Constraints*

The gradient ascent method has been shown to be able to efficiently generate new testing cases that maximize both Objective 1 and 2. However, not all testing cases generated by this method make sense in reality. Certain constraints have to be implemented according to different domain knowledge. We propose a complete set of image transformations treatment, including lighting effects, occlusions, and image distortions due to linear transformations, affine transformations, convolutional transformations, and composite transformations. We can use these transformations to cover more neurons in the DNNs. We also utilized the Generative Adversarial Networks (GAN) technique to further improve our capability to generate testing samples beyond the domain-specific image manipulations. By using GAN, we can generate images in an unsupervised way, transferring environment styles between two datasets, and generating images which cannot be easily obtained in the real image datasets. This capability has its realistic usage since some corner cases cannot be easily obtained in the real image datasets (extreme weathers, first-person-view collision scenes, etc.) This technique enables our testing system to overcome some limitations in our baseline system, as also mentioned in Ref. [4] [9]. The GANs can help us to generate a series of high-quality datasets by simulating a specific environment driven by desired events, with much lower cost.

Other special care has also been taken during the generation of new corner cases. For example, in image testing cases, if the pixel RGB value exceeds 0-255 bounds, the test case has to be rejected since it is not possible to generate such an image. These constraints or image transformations have simulated realistic image effects and led to meaningful erroneous inputs.

IV. EXPERIMENTAL SETUP

In this section, we will introduce the experimental setup for the project, including related hardware, software, datasets, and corresponding DNNs.

*A. Palmetto Cluster*

We have built the framework on the Palmetto Cluster. Palmetto is Clemson University's high performance computing (HPC) resource. Specifically, we ran our project work on Palmetto with the following specification:

- 1 CPU with 8 cores
- 64 GB RAM
- 2 GPU with GPU model as NVIDIA V100 on NVIDIA DGX-2 supercomputer cluster

*B. Datasets and DNNs*

In this project, we have focused our attentions on the image datasets, including MNIST, ImageNet, Driving dataset (Udacity self-driving car challenge dataset), and GTRSB dataset (the German traffic sign benchmark dataset).

For each of the first three datasets, we used three pre-trained DNNs: LeNet-1, LeNet-4, and LeNet-5 for MNIST; VGG-16, VGG-19, and ResNet50 for ImageNet; and DAVE-orig, DAVE-norminit, and DAVE-dropout for Driving. For the

GTRSB dataset, we have trained three DNNs (LeNet-1, LeNet-4, and LeNet-5). After obtaining the error-inducing samples, we augmented the training set to retrain the three models to evaluate the improvement in accuracy.

## V. BASELINE COMPARISON

We first deployed the DeepXplore framework [1] on our Palmetto system as a baseline. During this reimplementation process, we systematically studied the choices of hyperparameters for different DNNs. These hyperparameters represent the non-deterministic nature of the original algorithm. The hyperparameters $\lambda_1$ and $\lambda_2$ in Equation (3) are important parameters which were used to balance output differences of DNNs and balance coverage and differential behavior, respectively. The authors considered the optimization of them by fixing one parameter and changing the other one through a series of experiments. Here we explored a more comprehensive evaluation combing $\lambda_1$ and $\lambda_2$ together for optimal performance for our dataset. In addition, we explored the relationship between the different activation thresholds ($t$) with different DNNs.

| | | λ1 for the MNIST dataset | | | | | | |
|---|---|---|---|---|---|---|---|---|
| | | 0.5 | 1.0 | 1.5 | 2.0 | 2.5 | 3.0 | 3.5 |
| λ2 for the MNIST dataset | 0.5 | 338.5 | 325.9 | 340.8 | 342.4 | 352.1 | 357.4 | 343.7 |
| | 1.0 | 313.5 | 352.0 | 356.8 | 337 | 322.8 | 333.8 | 357.6 |
| | 1.5 | 349.9 | 336.6 | 336.7 | 325.3 | 347.5 | 357.3 | 346.9 |
| | 2.0 | 334.9 | 338.8 | 346.5 | 311 | 307.7 | 313.5 | 345.1 |
| | 2.5 | 349.7 | 339.6 | 339.6 | 345.2 | 331.1 | 342.3 | 345.5 |
| | 3.0 | 345.8 | 351.1 | 343.6 | 357.2 | 342.4 | 335.3 | 336.6 |
| | 3.5 | 347.9 | 342.1 | 365.2 | 347.0 | 352.6 | 351.4 | 338.5 |

*Table 1: Variation of runtime with different λ1 and λ2*

Table 1 shows the variation in the testing system runtime (in milliseconds) while generating the first difference-inducing input for the tested DNNs with different combinations of $\lambda_1$ and $\lambda_2$. In this experiment, we kept the step length as 10 and the threshold as 0. We ran each combination over 10 times and computed the average runtime without extremums. In this table, the red value represents the fastest runtime when we hold λ2 and gradually change λ1. Similarly, the green value represents the fastest runtime, while λ1 is held and λ2 is adjusted. We found optimal combinations only achieved when λ1 and λ2 are equal or close, meaning that we cannot maximize differential behaviors or maximize neuron coverage alone. In this project, we choose the optimal combination of $\lambda_1$ and $\lambda_2$ as 2.5 and 2.0, respectively.

| | MNIST/ImageNet with λ1 = 2.5, λ2 = 2.0, step = 1, seeds = 10 | | | | | | |
|---|---|---|---|---|---|---|---|
| t | 0.001 | 0.01 | 0.05 | 0.1 | 0.5 | 0.99 | 1 |
| Average neuron coverage of MNIST | 95.1% | 92.7% | 90.4% | 78.8% | 40.2% | 11.1% | 0.00% |
| Average neuron coverage of ImageNet | 95.7% | 91.6% | 80.7% | 69.1% | 34.0% | 1% | 0.00% |

*Table 2: The variation in average neuron coverage*

The thresholds ($t$) is a critical hyperparameter to determine whether the current neuron is activated. Table 2 describes the variation in average neuron coverage of MNIST and ImageNet datasets while generating 10 error-inducing inputs for the tested DNNs. Generally, the value of threshold should belong to the interval from 0 to 1. According to Table 2, no matter which dataset we chose, the average neuron coverage followed an apparent decrease with the threshold increasing. In addition, under the same threshold value, we can find the average neuron coverage of ImageNet was commonly lower than that of MNIST. This phenomenon matches the fact that the DNNs for ImageNet is usually much more complex than that for MNIST, thus more difficult for us to cover more neurons.

## VI. DIFFERENTIAL BEHAVIOR BASED AUGMENTING TRAINING

In this project, we explored one of the applications of the automatically generated corner case samples as augmenting training. Similar to the adversarial training, the augmenting training augments the original training set with error-inducing samples to fix the error behaviors and eventually increase the robustness and prediction accuracy of the DNNs.

In this project, we chose to work on the GTRSB dataset on the traffic sign recognition task. Recent work has shown that adversarial stickers for Stop signs, which mimic the graffitis on road signs, can be developed to fool the standard road sign classification neural networks [3]. Specifically, the Stop signs with such adversarial stickers could be classified as speed limit signs under various angles and lighting conditions. We used the proposed testing system to generate similar error-inducing samples by adding occlusion to the original images, either with a single gradient directed rectangle or multiple white/black dots. Figure 1 shows the real graffiti on a Stop sign, the adversarial stickers on a Stop sign, and occlusion contained images generated by this project, respectively.

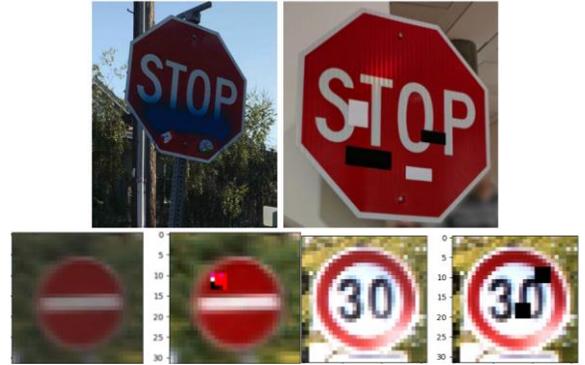

*Figure 1: top left is the real graffiti on a Stop sign; top right is the image of adversarial stickers on a Stop sign; bottom images show different occlusions on the images of road signs*

To generate augmenting training samples, we need to train three similar DNNs for the GTRSB dataset, in which case we chose LeNet-1, LeNet-4, and LeNet-5. LeNet-1 is the simplest DNN among the three, with only one convolutional layer and one max-pooling layer and without any hidden dense layer;

while LeNet-5 has two convolutional layers, two max-pooling layers and two hidden dense layers. After we trained the three DNNs with the 39,209 training images in the GTRSB dataset, we tested them with the 12,630 testing images, achieving testing accuracy as 88.6%, 90.7%, and 91.8%, respectively. The accuracies for three models were reasonably high for the standard testing images but will drop significantly with the addition of special attacking samples, as will be seen in the next paragraph.

Then, we randomly selected 100 images from the testing dataset as the seeds and fed them to the automated testing system framework. The testing system identified whether the seed images already generated different behaviors between the test DNNs. If not, it would generate new samples by adding occlusion to the original images. Both already differed original images and newly generated images were added together to form an augmented dataset, which were used for further augmenting training. We achieved more than 90% conversion from the seed images to the augmented dataset. It is worth to note that we didn't label the generated augmenting data with the majority voting between the three DNNs, as we noticed that such majority voting was not always correct. On the other hand, we recorded the original label from the testing set, and passed it to the augmenting data whenever necessary.

During the generation of augmented dataset, we noticed the neural coverage increasing from 48.3% for 10 processed seeds to 68.2% for 100 processed seeds. We further increased the number of seeds to 1000 and increased the neural coverage to 78.5% at the end of sample generation. A detailed study for neural coverage of GTRSB dataset is shown in Table 3.

| Seeds | 10 | 20 | 30 | 100 | 300 | 400 | 1000 |
|---|---|---|---|---|---|---|---|
| Neuron Coverage | 53.4% | 58.3% | 61.4% | 66.1% | 75.5% | 76.3% | 78.5 |

*Table 3: The variation in average neuron coverage with different number of seeds for occlusion with black dots*

Before augmenting training, the accuracy of three DNNs on the augmenting data is relatively low. For example, for occlusion images generated from 1000 seeds, the accuracies were only 23.8%, 58.6%, and 57.1%, respectively. To perform the augmenting training, we combined the augmenting dataset with the original training dataset to form a new training set. Then, we retrained the three DNNs with the latest training set. To validate the effectiveness of the differential behavior-based augmenting training, we also performed two other augmenting training approaches for comparison. The first control approach involved generating an augmenting dataset with the same amount of unchanged images. Each image in the augmenting set was randomly selected from the original testing set. In the second control approach, a similar augmenting set was constructed except that each image inside was randomly generated from the 'ImageGenerator' of the Keras framework using random original training data as seeds.

The comparison of three augmenting training approaches is shown in Figure 2. Without any augmenting training, all three DNNs shows low accuracy for the augmenting dataset. The accuracy for LeNet-1 was especially low since it had the simplest structure compared to the other two. When the training set was augmented with the augmenting dataset, we retrained the three DNN models. After retraining, the accuracies had been improved dramatically for all three models, achieving 85.5%, 90.0%, and 90.4%, respectively.

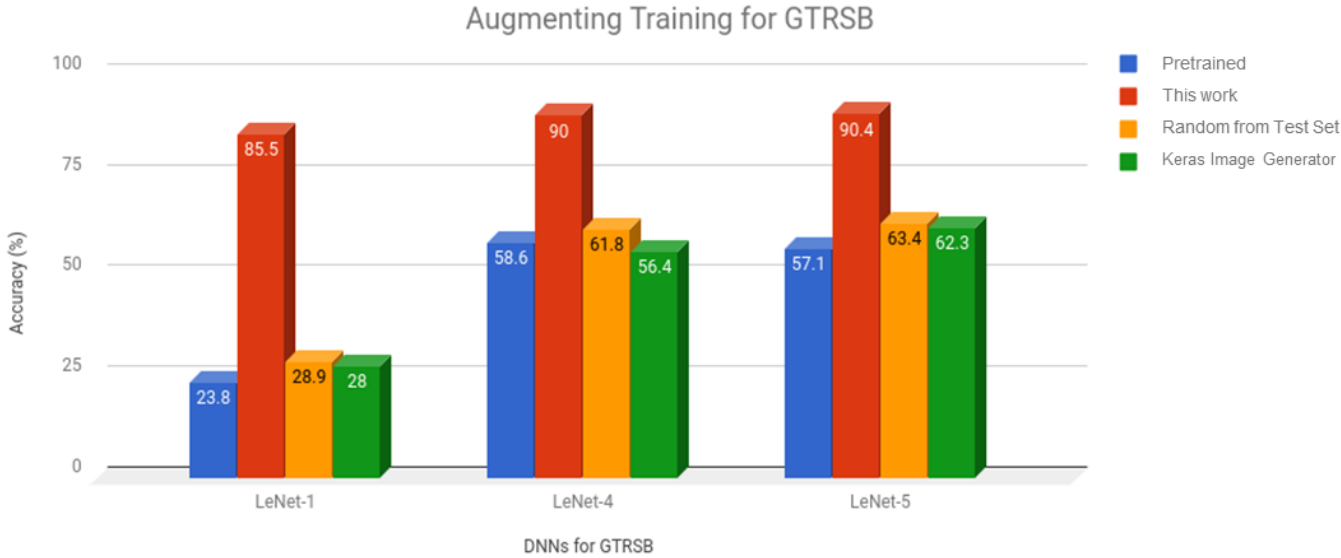

*Figure 2: Comparison of three augmenting training approaches in terms of accuracy of three LeNet DNNs*

The improvements for the accuracies of the three DNNs are evident: 259.2%, 53.6%, and 58.3%, respectively. The new accuracies were now close to the ones for the original testing set. On the other hand, two randomly augmented data sets in the control groups only exhibited a slight increase in accuracy. The result shows that augmenting training with the proposed testing scheme can effectively fix the erroneous behaviors caused by the corner cases and thus significantly increase the robustness of the corresponding DNNs. In contrast, the other randomly augmenting training cannot find related corner cases, therefore, failed to fix the problem. While the result shows that the random transformation added by the ImageGenerator cannot cover the occlusion corner case, we later discovered that both random augmenting training helped to fix corner cases for affine transformation, although not as significant as proposed differential-behavior-based samples.

## VII. IMAGE TRANSFORMATION

In this part of the project, we further propose a complete set of image transformations. Two reference papers have discussed the important roles of image transformation in automated testing [4, 5]. It has been shown that different image transformations may activate different neurons in the neural networks [4]. Therefore, incorporating more image transformations and possibly combining them into the automated testing system could potentially result in generating more diverse erroneous inputs for automated testing of the DL systems.

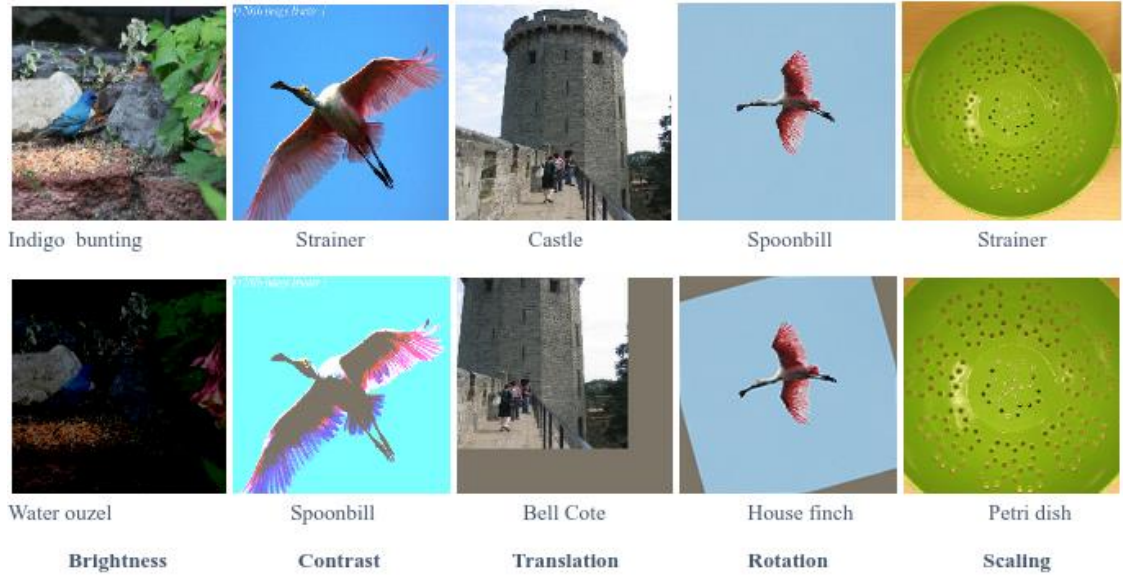

*Figure 3: upper row shows the original images with correct labels;  
bottom row shows the misclassified images generated by linear or affine transformations with incorrect labels*

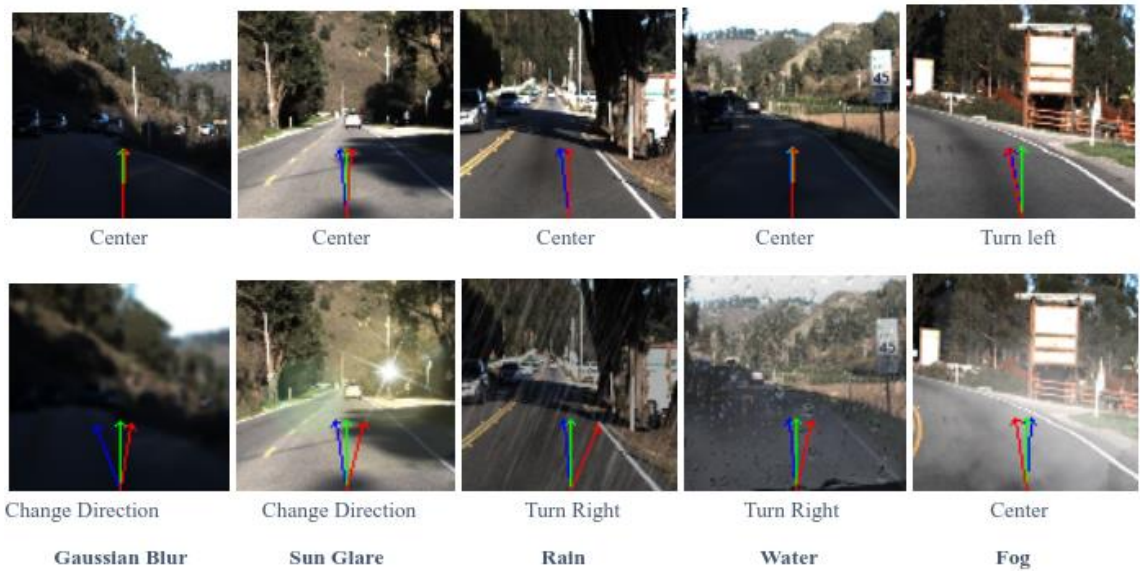

*Figure 4: Sample results of differential behavior of three state-of-art deep learning models by adding realistic driving conditions*

The image transformations on seed images need to be realistic and representative of real-world conditions like object movements or weather conditions. We decided to implement four groups of image transformations in this part of the project, including linear transformations such as changing contrast and lightness; affine transformations like translation, rotation, and scaling; convolutional transformations such as Gaussian blurring; and composite transformations, e.g., adding different weather conditions to the seed images. We now describe the details of these transformations below.

Changing brightness and contrast were implemented by using two linear transformations (Figure 2). Specifically, brightness corresponded to the bias term in the following equation, while contrast corresponded to the gain term. When we adjusted the brightness of an image, we just needed to add or subtract a constant value to each pixel's current bias value. In the automated testing system, we used the gradient of loss functions to determine such constant so that we can generate error-inducing samples with the least transformations in an unsupervised way. Similarly, the contrast was based on the difference between the gain values of different pixels, so we can adjust the contrast of an image by multiplying a constant value for each pixel.

$$Out(RGB, c) = In(RGB, c) * gain + bias$$

Affine transformations include translation, rotation, and scaling (Figure 2). We used a linear mapping method that preserved points, straight lines, and planes. Sets of parallel lines remain parallel after an affine transformation. Generally, the affine transformations are implemented as a matrix multiplication between a 2D image matrix and a 2 by 3 transformation matrix. Typical transformation matrices are shown in Figure 5. In our project, we again used the gradient of the loss function to determine the displacement for translation, the scaling factor, or the angle of rotation. By doing this way, we achieved the best transformation effect in a small number of iterations.

$$\begin{bmatrix} 1 & 0 & t_x \\ 0 & 1 & t_y \end{bmatrix} \begin{bmatrix} \alpha & \beta & (1-\alpha) \cdot center.x - \beta \cdot center.y \\ -\beta & \alpha & \beta \cdot center.x + (1-\alpha) \cdot center.y \end{bmatrix}$$

*Figure 5: left is the transform matrix for translation; right is the transform matrix for scaling combined with rotation*

Blurring is a convolutional transformation, which can be realized by adding each pixel of the input image with different transform-specific kernels to its local neighbors. In addition, the convolutional transformation included several different blurring filters. In our project, we chose Gaussian-blur with 3-by-3 kernel size to realize the convolutional transformation. Sigma, an important parameter of the Gaussian-blur filter, can be used to optimize the discrete approximation of the continuous Gaussian kernel. In our implementation, we used the product of the mean of gradients and the step length to determine the value of parameter sigma.

The primary purpose of composite transformation was to add different realistic driving conditions for the input images. To be specific, we introduced the following four artificial effects to simulate adverse weather conditions on the road: sun glare, rain, water drops on the windshield, and fog. We again used the concept of gradient ascending method to generate a new image. For instance, for the sun glare effect, we created two masks for the original and overlay sun glare effect. We closely monitored the gradients change from the first pass and only applied non-zero transparency values in the region that the differential gradient was more significant. The gradient calculated by TensorFlow was used to change the weight hyperparameter between the background and overlay images. Again, we recorded the image that generated the most significant differential behavior between tested DL models and marked the new image as a unique corner case. A set of sample result with differential predictions are shown in Figure 4.

### VIII. GAN-BASED AUTOMATED TESTING SAMPLE GENERATION

After evaluating new image transformation techniques, we observed that it was necessary to improve the quality of the testing samples to reflect the real world adverse conditions. The transformed images from either our proposed methods or state-of-art sample generators in Keras were often not realistic. They had little chance to be encountered in the real-world application. The Generative Adversarial Networks (GAN) method, an emerging deep learning technique, was then selected to generate photo-realistic testing samples automatically.

The schematic working mechanism of the GAN is shown in Figure 6. In this method, the input will be the original testing samples (MNIST handwritten digits or photos from Udacity driving dataset). A generic GAN system consists of two basic models: a generator (G) and a discriminator (D). Initially, the system will introduce a certain artificial noise to the training samples and pass it through the generator. The generator will then try to generate a new image that closely mimics the original testing samples. The generated sample will then be sent to the discriminator. The discriminator model is a pre-trained (and will be continuously trained) classifier for determining whether a given image data set looks like a real image or artificially created image. This is basically a binary classifier; it usually takes the form of a convolutional neural network (CNN). The classification result will then be fed back to the generator through gradient descent. The generator and discriminator will then learn from the past results and improve the accuracy in generating or discriminating the testing samples. The end result of such a deep learning system is that the generator can learn to generate new testing samples that closely resembles the original samples. In contrast, the discriminator learns how to find real and fake images more accurately. There are many variations based on the GAN concept, such as Deep Convolutional GAN (DCGAN) and CycleGAN. We used both of these two methods for different applications.

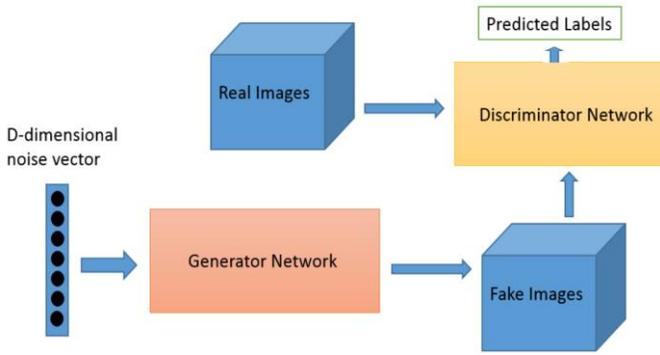

*Figure 6: Schematics of Generative Adversarial Networks*

*A. GAN applied to MNIST Dataset*

The Deep Convolutional Generative Adversarial Networks (DCGAN) [7] was used to train the testing system to automatically generate handwriting digits through learning from the MNIST training samples. The MNIST dataset provided many grayscale images (28x28 pixel) of handwritings for each digits, which was fed into the DCGAN generator together with a D-dimensional noise vector (784x1) as shown in Figure 9. The generator of DCGAN was based on a convolutional network so that the end result was the fusion between the real image and random noise. At each iteration, the generated fake image was classified using the discriminator with comparison to real images. At the end of each iteration, the classification results were passed to the next iteration by backpropagation.

*B. GAN applied to Driving Dataset*

The generated MNIST testing samples are representative of how computer AI learns to generate images following human handwriting styles. However, for other more complicated scenarios, such as the autonomous car driving dataset, it is not realistic to train the GAN to learn from scratch. A more feasible way for such a situation is to transfer an existing dataset into the same style of the given training dataset. In the application of the autonomous driving dataset, we proposed to make use of virtual images generated from a car driving simulator. We attempted to generate photo-realistic images that can inherit styles from the given car driving dataset. The motivation of this proposed work is that the simulator can be much more flexible in simulating different testing environments and generating a large amount of testing samples in a short amount of time. Many artificial feature change is made possible through this way, such as daytime to nighttime environment change, season or light change to ambient objects (lane, road condition, etc.)

The proposed style transferring task is a strong forte of GAN compared to other deep learning networks. Therefore, we implemented a more sophisticated GAN model called CycleGAN [8] for this purpose. In CycleGAN, we provided the model two training sets: train_A and train_B. Also, in addition to the model G (generator) and D (discriminator), the new system added an inverse mapping model (F). During training, the CycleGAN system will randomly pick one image from each of the two training sets. The training goal is to transfer the style of the training set A to image X (from B) and, at the same time, transfer style of training set B to Y (from A). The inverse mapping (F) provides a means to ensure the transformation is backward reversible. A schematic of this CycleGAN is shown in the figure below. The advantage of the CycleGAN is that each pair of training images is randomly picked from the dataset folders. Therefore the training can be completely unsupervised. This method is ideal for our automated testing sample generator task.

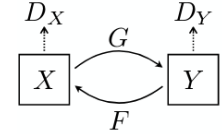

*Figure 7: Schematics of CycleGAN*

A sample output from CycleGAN running on the Pytorch framework is shown in the figure below. At each epoch, the loss functions from G, D, F models for both testing samples were minimized through a fixed 5000 iterations. This sample result is a representative result at epoch number 9. In the original paper, the authors showed that successful results could be obtained only based on 5 epochs. A detailed zoom-in comparison can be found in the figure below:

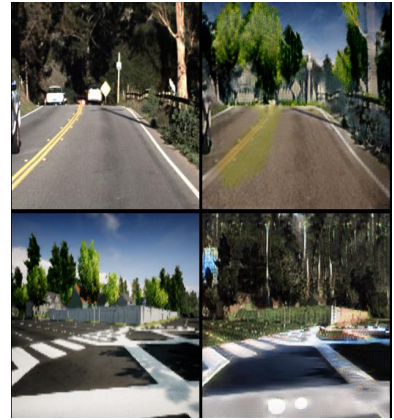

*Figure 8: Example testing images generated by GAN: style transferring between real and virtual images*

The above image demonstrates the capability of CycleGAN in transferring the ambient environment between the virtual and real image datasets. The images from the first column are the original input images, whereas the second column contains the generated "fake" images based on the other dataset. The first row uses the real world image from the car driving dataset, whereas the second row uses the computer simulated car driving images as input. In this way, we transferred the real image into a virtual environment style and vice-versa, the virtual image into a realistic environment. It can be clearly seen that the sky in both real and virtual images was successfully identified and swapped during this style transferring process.

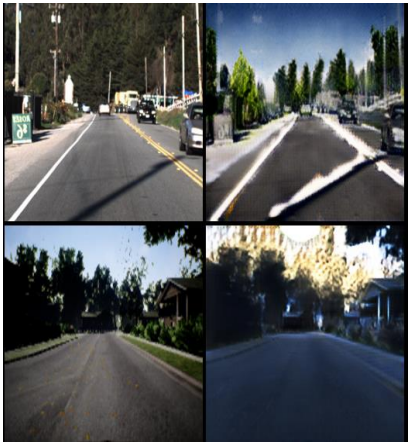

*Figure 9: Example testing images generated by GAN: modifying background (sky, trees)*

In another example, it can be seen that the road condition and lighting effects have also been adjusted accordingly (for example, the bottom row in Figure 9). In this way, we validated the capability of the automated sample generation capability using the GAN technique. The additional advantage of introducing this technique is that it can generate a large amount of training data for specific environments. The visual testing samples often take time to collect and often have difficulties in obtaining data from extreme conditions. Using this proposed method, we can take the existing available dataset and transfer it into the desired environment without losing detailed features. The deep learning system under testing should be inert to the artificial change to the testing samples (for example, an autonomous driving system should behave the same if tested under a simulated nighttime driving dataset that was initially transferred from recorded daytime image sets). This method can be used as an event-driven testing sample generation model and have a high potential to explore the extreme corner cases for deep learning systems.

## IX. CONCLUSION

In this project, we proposed an automated testing framework for deep learning systems based on differential behavior criteria. During our implementation, we studied the optimization of the hyperparameters of the framework. We introduced new augmented training data with error-inducing samples to improve robustness against erroneous behaviors and malicious inputs. By retraining the model based on automated generated corner cases, the accuracy of three generic models on a traffic sign recognition dataset increased by 259.2%, 53.6%, and 58.3%, respectively. We have also proposed a complete set of image transformations to support linear transformations, affine transformations, convolutional transformations, and composite transformations. Finally, we used GAN technique to further improve our capability to generate testing samples.

## X. ACKNOWLEDGMENTS

This work was supported in part by ONR award N00014-19-1-2295

REFERENCES

[1] Pei, K., Cao, Y., Yang, J., & Jana, S. (2017, October). Deepxplore: Automated whitebox testing of deep learning systems. In Proceedings of the 26th Symposium on Operating Systems Principles (pp. 1-18). ACM.

[2] Yosinski, J., Clune, J., Nguyen, A., Fuchs, T., & Lipson, H. (2015). Understanding neural networks through deep visualization. arXiv preprint arXiv:1506.06579.

[3] Ivan Evtimov, Kevin Eykholt, Earlence Fernandes, Tadayoshi Kohno, Bo Li, Atul Prakash, Amir Rahmati, and Dawn Song. Robust physical-world attacks on machine learning models. arXiv preprint arXiv:1707.08945, 2017.

[4] Tian Y., Pei, K., Jana, S. & Ray, B. (2017). DeepTest: Automated Testing of Deep-Neural-Network-driven Autonomous Cars. arXiv preprint arXiv:1708.08559.

[5] Pei, K., Cao, Y., Yang, J., & Jana, S. (2017). Towards Practical Verification of Machine Learning: The Case of Computer Vision Systems. arXiv preprint arXiv:1712.01785.

[6] Goodfellow, I., Pouget-Abadie, J., Mirza, M., Xu, B., Warde-Farley, D., Ozair, S., ... & Bengio, Y. (2014). Generative adversarial nets. In *Advances in neural information processing systems* (pp. 2672-2680).

[7] Radford, A., Metz, L., & Chintala, S. (2015). Unsupervised representation learning with deep convolutional generative adversarial networks. arXiv preprint arXiv:1511.06434.

[8] Zhu, J. Y., Park, T., Isola, P., & Efros, A. A. (2017). Unpaired image-to-image translation using cycle-consistent adversarial networks. *arXiv preprint arXiv:1703.10593*.

[9] Zhang, M., Zhang, Y., Zhang, L., Liu, C., & Khurshid, S. (2018). DeepRoad: GAN-based Metamorphic Autonomous Driving System Testing. arXiv preprint arXiv:1802.02295.